\documentclass[11pt]{article}
\usepackage[margin=1.0in]{geometry}
\usepackage{cite}
\usepackage{amsmath,amssymb,amsfonts}
\usepackage{graphicx}
\usepackage{textcomp}
\usepackage{booktabs}
\usepackage{multirow}
\usepackage{algorithm}
\usepackage{soul}
\usepackage{subfig}
\usepackage{hyperref}
\usepackage{bbm}
\usepackage{pifont}
\usepackage{url}
\hypersetup{hidelinks}
\newcommand{\zb}[1]{#1}
\def\BibTeX{{\rm B\kern-.05em{\sc i\kern-.025em b}\kern-.08em
    T\kern-.1667em\lower.7ex\hbox{E}\kern-.125emX}}

\title{Morphology-driven single-domain generalization for cross-cohort Alzheimer's disease detection using 3D MRIs}

\author{
Zobia Batool$^{1}$, Diala Lteif$^{2}$, Vijaya B. Kolachalama$^{3}$,\\
Huseyin Ozkan$^{1}$, and Erchan Aptoula$^{1}$
}

\date{}

\begin{document}
\maketitle

\begingroup
\renewcommand{\thefootnote}{}
\footnotetext{
$^1$Faculty of Engineering and Natural Sciences, Sabanci University, Istanbul, T\"urkiye.
$^2$Department of Computer Science, Boston University, Boston, MA, USA.
$^3$Department of Medicine, Computer Science, and Computing and Data Sciences, Boston University, Boston, MA, USA.
Correspondence: zobia.batool@sabanciuniv.edu; dlteif@bu.edu; vkola@bu.edu.
}
\endgroup

\maketitle
\begin{abstract}
Structural MRI is commonly used to detect brain changes associated with cognitive impairment, including those related to Alzheimer’s disease (AD). However, deep learning models often generalize poorly across datasets due to class imbalance, scanner variability, and demographic differences. We present a morphology-driven single domain generalization (SDG) framework that trains on a single source domain and generalizes to unseen target domains without retraining. The method combines learnable class-conditioned morphological augmentations with supervised contrastive learning to increase intra-class variation and encourage discriminative representations. Trained on structural MRI from the National Alzheimer’s Coordinating Center (NACC; n=4,647) and evaluated on three independent cohorts (n=3,126), the proposed method numerically improves mean F1 by 5.0 percentage points over the strongest competing method across the external cohorts, despite class imbalance and acquisition heterogeneity. Source code will be made available upon acceptance at \url{https://github.com/zobia111/MorphoSDG}.
\end{abstract}

\paragraph{Keywords:} Alzheimer's disease, dementia, cognitive impairment, domain generalization, mathematical morphology, contrastive learning, domain shift.

\section{Introduction}
\label{sec:introduction}

Structural magnetic resonance imaging (sMRI) is widely used to assess structural brain changes associated with cognitive impairment, including Alzheimer’s disease (AD). Regional atrophy and other structural abnormalities can provide supportive evidence of neurodegeneration, motivating the use of machine learning (ML) methods to associate sMRI-derived patterns with diagnostic categories. However, models trained on sMRI often generalize poorly across datasets because of class imbalance, scanner and protocol variability, and demographic differences \cite{heidari2026bi, liu2026revisiting}. These limitations undermine out-of-distribution performance and reduce reliability in real-world deployment.

To address these limitations, recent studies have proposed a range of deep learning architectures. Convolutional neural networks (CNNs) have been enhanced with spatial and channel attention mechanisms \cite{li2024multi}, while vision transformers (ViTs) use self-attention to model long-range dependencies and multimodal information \cite{joy2025vitad, duong2025multimodal, kunanbayev2024training}. Hybrid CNN-attention architectures \cite{gan2025dense, jiang2024anatomy} and graph-based models \cite{gamgam2024disentangled, zhou2024multi} have also been investigated. In parallel, domain-aware learning and harmonization methods aim to reduce inter-site variability and align feature distributions across cohorts \cite{cai2023prototype, scholz2025contrastive}. However, some of these approaches require access to target-domain data or additional adaptation, limiting their applicability when retraining or target data are unavailable. This motivates single domain generalization (SDG), where a model is trained on a single accessible source dataset and evaluated directly on unseen domains, with the aim of improving robustness across imaging protocols and populations.

As far as the realistic case of non-matching distributions between training and testing sets is concerned, various strategies have been proposed to address domain shift in AD classification, including attention-guided models using biologically informed priors \cite{Diela}, prototype-guided multi-scale domain adaptation \cite{cai2023prototype}, domain-knowledge-constrained models \cite{sMRI}, and augmentation strategies such as DT-Mixup \cite{batool2025distance}. \zb{Although these methods address distributional variability through different mechanisms, we investigate learnable 3D morphological transformations combined with supervised contrastive learning for cross-cohort classification.} Moreover, models relying primarily on global structural representations may generalize poorly across external cohorts \cite{dong2025convolutional}, while dataset-level harmonization may not fully preserve subject-specific anatomical variation \cite{scholz2025contrastive}. Therefore, existing domain generalization (DG) strategies for AD classification, although conceptually diverse, do not explicitly model class-associated structural variation through learnable morphological augmentation during training.

In this article, we propose an end-to-end morphology-driven SDG framework for AD classification. The method uses the shape-processing properties of mathematical morphology \cite{serra2023mathematical} to generate learnable class-conditioned 3D augmentations and combines them with supervised contrastive learning to increase intra-class variation and encourage robust representations. The framework is built on a 3D U-Net encoder \cite{Models_Genesis} and is trained on NACC \cite{NACC} as a single source cohort without access to target-domain data. It is evaluated on three unseen target cohorts, ADNI \cite{ADNI}, AIBL \cite{AIBL}, and OASIS-3 \cite{OASIS}, to assess generalization across heterogeneous imaging datasets.

\section{Related Work}
\subsection{AD Classification Approaches}
Numerous studies have investigated AD classification using MRI data. In detail, a Multi-Attention-based Global 3D ResNet architecture was introduced in \cite{li2024multi}, using channel and spatial attention together with a non-local block to improve feature representation and capture long-range dependencies on ADNI. Similarly, the Dense Attention Network \cite{gan2025dense} combined convolutional layers with linear attention for strong classification and a low parameter count. Another study proposed an ensemble framework \cite{rahim2025early} to combine CNN-based feature extraction with attention modules and Bayesian-optimized classifiers for early AD prediction from multi-view 3D MRI. 

In the context of ViTs, the Joint Transformer architecture \cite{alp2024joint} extracted features from 3D MRI scans modeled as sequences to capture spatial interdependencies. These features were then classified using a time-series transformer to learn both spatial and progression-related patterns for AD classification. Alternatively, the multi-modal surface-based transformer model \cite{duong2025multimodal} integrated MRI and PET scans using a mid-fusion design with both self-attention and cross-attention blocks, outperforming volume-based baselines. A modified ViT model in \cite{joy2025vitad} was trained on preprocessed MRI slices with patch embedding and multi-head self-attention for 5-stage AD classification. Finally, \cite{lu2025efficient} introduced RanCom-ViT, which enhanced classification efficiency and accuracy through token compression and a random vector functional-link structure in the classification head.

As far as graph-based neural networks are concerned, a prompted hypergraph neural network captured higher-order multimodal relationships while using hypergraph prompt learning to adapt the model efficiently with limited data. This approach reduced the number of trainable parameters while improving DG for AD diagnosis and MCI conversion prediction \cite{liu2026diagnosing}. Additionally, a hybrid architecture called 3D HCCT \cite{majee2024enhancing} integrated 3D convolution and ViTs with hybrid pooling and an automated preprocessing pipeline for efficient, end-to-end AD classification using 3D MRI.

\subsection{Morphological Modules for AD Classification}
Mathematical morphology has been extensively explored in the context of medical imaging \cite{james2014medical} before the advent of deep learning due to its inherent capacity for shape analysis \cite{aptoula2009multivariate}. Consequently, multiple attempts have been made to implement its data-driven counterpart, even though its non-linearity complicates differentiability and the back-propagation of errors via gradient descent. For instance, one approach \cite{hu2022learning} introduced differentiable dilation and erosion using neural architecture search to integrate them into end-to-end models. Another method \cite{ghosh2024multi} proposed multi-scale modules with learnable structuring elements, and a separate effort \cite{guzzi2024differentiable} developed soft morphological filters to optimize operations like dilation. A hybrid strategy \cite{canales2023hybrid} combined fixed morphological layers with shallow convnets, though it lacked flexibility and full end-to-end training.

More specifically, in the context of morphology-oriented AD classification, graph-theoretical metrics were applied to morphological networks for early classification \cite{ding2023topological}, while mesh-based graph convolutional networks were used in \cite{azcona2020interpretation} to model cortical structure. \zb{Finally, a transformer-based model \cite{medtransform} used class-dependent 2D erosion and dilation with a predefined, nonlearnable structuring element to augment MRI slices. By comparison, our proposed approach employs fully differentiable 3D morphological modules with learnable class-specific structuring elements and sharpness parameters, using the resulting volumes in supervised contrastive learning.}

\subsection{Domain Generalization in AD Classification}
In DG, training and testing domains have distinct distributions, and in the case of SDG, there is only one training domain. AD classification, especially under the SDG setting, is still relatively underexplored. A disease-informed framework based on a two-stage 3D U-Net pipeline, guided by saliency-based attention to improve generalization, was introduced in \cite{Diela}. Another approach \cite{sMRI} used a patch-free ResNet architecture with domain-specific encoding to separate invariant and variant features. In contrast, Prototype-guided multi-scale domain adaptation \cite{cai2023prototype} addressed both marginal and conditional distribution shifts through a mix of adversarial and metric-based alignment modules. Similarly, a recent multisite MRI study combined a 3D CNN with a hybrid spatial-channel attention mechanism and contrastive learning to extract domain-invariant features for AD classification across heterogeneous imaging domains \cite{sam2026multisite}. Finally, a recent contribution \cite{batool2025distance} introduced a new augmentation strategy using distance transform-based mixing across MRI samples. \zb{However, existing DG methods do not leverage the anatomical changes realized in the brain due to AD, and rather constitute generic approaches. The proposed method aims to address these limitations by incorporating learnable morphological cues into the generalization process.}

\section{Proposed SDG Approach}
\begin{figure}[ht]
    \begin{center}
        \includegraphics[width=1.0\linewidth]{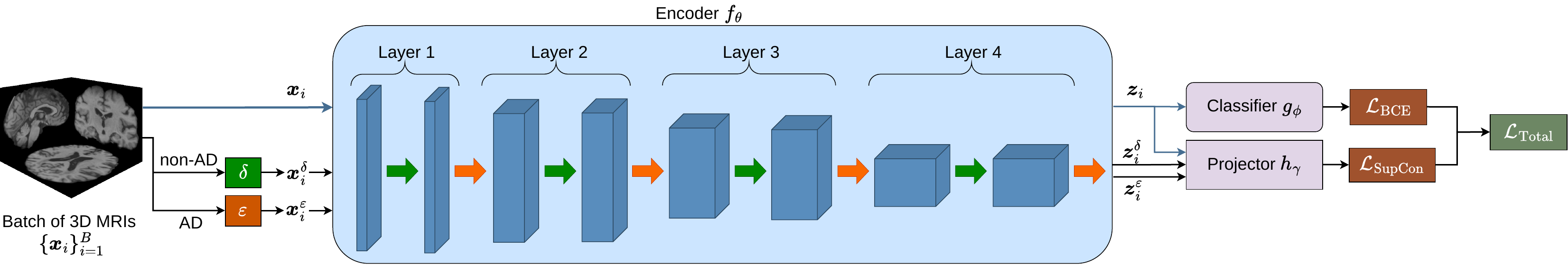}
        \caption{\textbf{Overview of the proposed training pipeline.} For each sample $\boldsymbol{x}_i$ in a batch $\{\boldsymbol{x}_i\}_{i=1}^{B}$, the encoder $f_{\theta}$ produces the original representation $\boldsymbol{z}_i=f_{\theta}(\boldsymbol{x}_i)$. This representation is fed into the classifier $g_\phi$ to compute the binary cross-entropy loss $\mathcal{L}_{\text{BCE}}$. In parallel, each sample $\boldsymbol{x}_i$ is routed to a class-conditional learnable morphological module according to its label: the pseudo-dilation module $\delta$ is applied to non-AD samples to generate the pseudo-dilated sample $\boldsymbol{x}_i^{\delta}$, while the pseudo-erosion module $\varepsilon$ is applied to AD samples to generate the pseudo-eroded sample $\boldsymbol{x}_i^{\varepsilon}$. The corresponding transformed image is then encoded to obtain either $\boldsymbol{z}_i^{\delta}$ or $\boldsymbol{z}_i^{\varepsilon}$. The original representation $\boldsymbol{z}_i$, together with its corresponding class-conditional transformed representation, is passed through the projection head $h_\gamma$ to compute the weighted supervised contrastive loss $\mathcal{L}_{\text{SupCon}}$. Finally, $\mathcal{L}_{\text{BCE}}$ and $\mathcal{L}_{\text{SupCon}}$ are combined to form the total training objective $\mathcal{L}_{\text{Total}}$.}
        \label{fig:training_pipeline}
    \end{center}
\end{figure}

The proposed method combines a 3D U-Net encoder with class-specific 3D learnable morphological augmentations and supervised contrastive learning in order to improve AD classification under a realistic SDG setting. This combination of components encourages the model to generate more diverse variations while simultaneously encouraging class-consistent representations. During training, for AD and non-AD samples, 3D learnable morphological augmentations are computed via pseudo-erosion and pseudo-dilation respectively, motivated by the fact that these operators can visually simulate local morphological changes associated with brain atrophy. Both original and augmented samples are processed through the encoder and a projection head to compute contrastive embeddings. The total loss combines weighted binary cross-entropy loss with inverse class frequency weights and supervised contrastive loss  as shown in Fig.~\ref{fig:training_pipeline}.

\subsection{Model Architecture}
The classification framework is built upon a 3D U-Net architecture \cite{Models_Genesis} as the backbone for feature extraction. To adapt the architecture for the task under study, the decoder is dropped, and the encoder is retained as a feature extractor as illustrated in Fig.~\ref{fig:training_pipeline}. The network comprises four sequential convolutional blocks, each containing two convolutional layers followed by batch normalization and ReLU activation functions. To incorporate prior knowledge, the model is initialized with pre-trained weights obtained from chest CT imaging data \cite{Models_Genesis}. The extracted high-level feature maps are globally average pooled, followed by two fully connected layers to generate the final classification output.

\subsection{Class-Specific Morphological Augmentation}
\label{sec:morphological_augmentation}

To address distributional shifts between training and testing domains, class-specific learnable 3D morphological augmentations are incorporated into the training pipeline. Standard augmentation techniques typically apply identical transformations regardless of class and may blur class-associated structural differences. To preserve these cues, the proposed method employs two learnable morphological modules based on differentiable morphology \cite{mondal2019morphological}: a pseudo-dilation module \(\delta\) and a pseudo-erosion module \(\varepsilon\).

\begin{figure}[ht]
  \begin{center}
  \subfloat[Original]{%
    \includegraphics[width=0.3\linewidth]{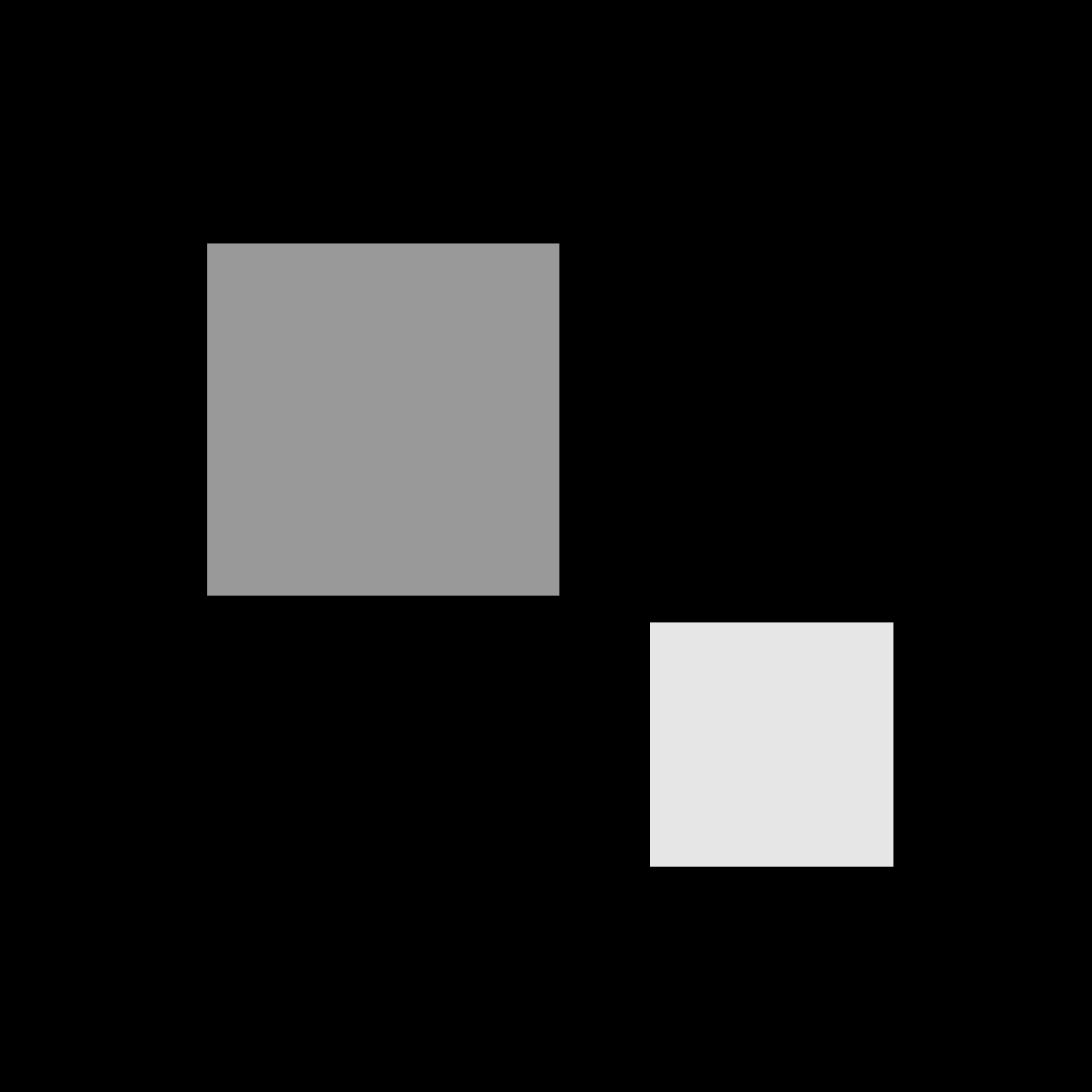}
    \label{fig:toy_original}%
  }
  \hfill
  \subfloat[Non-flat dilation]{%
    \includegraphics[width=0.3\linewidth]{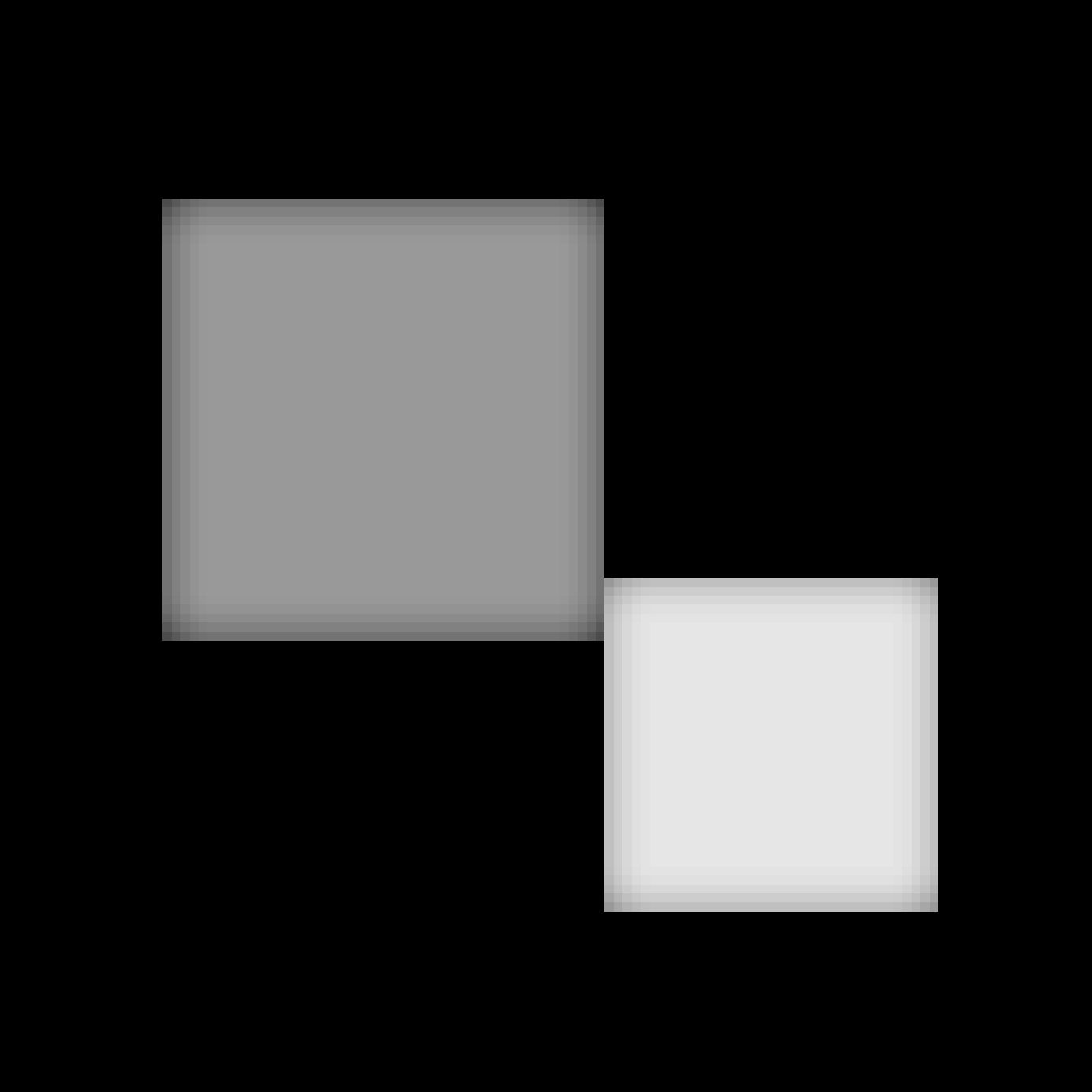}
    \label{fig:toy_dilation}%
  }
  \hfill
  \subfloat[Non-flat erosion]{%
    \includegraphics[width=0.3\linewidth]{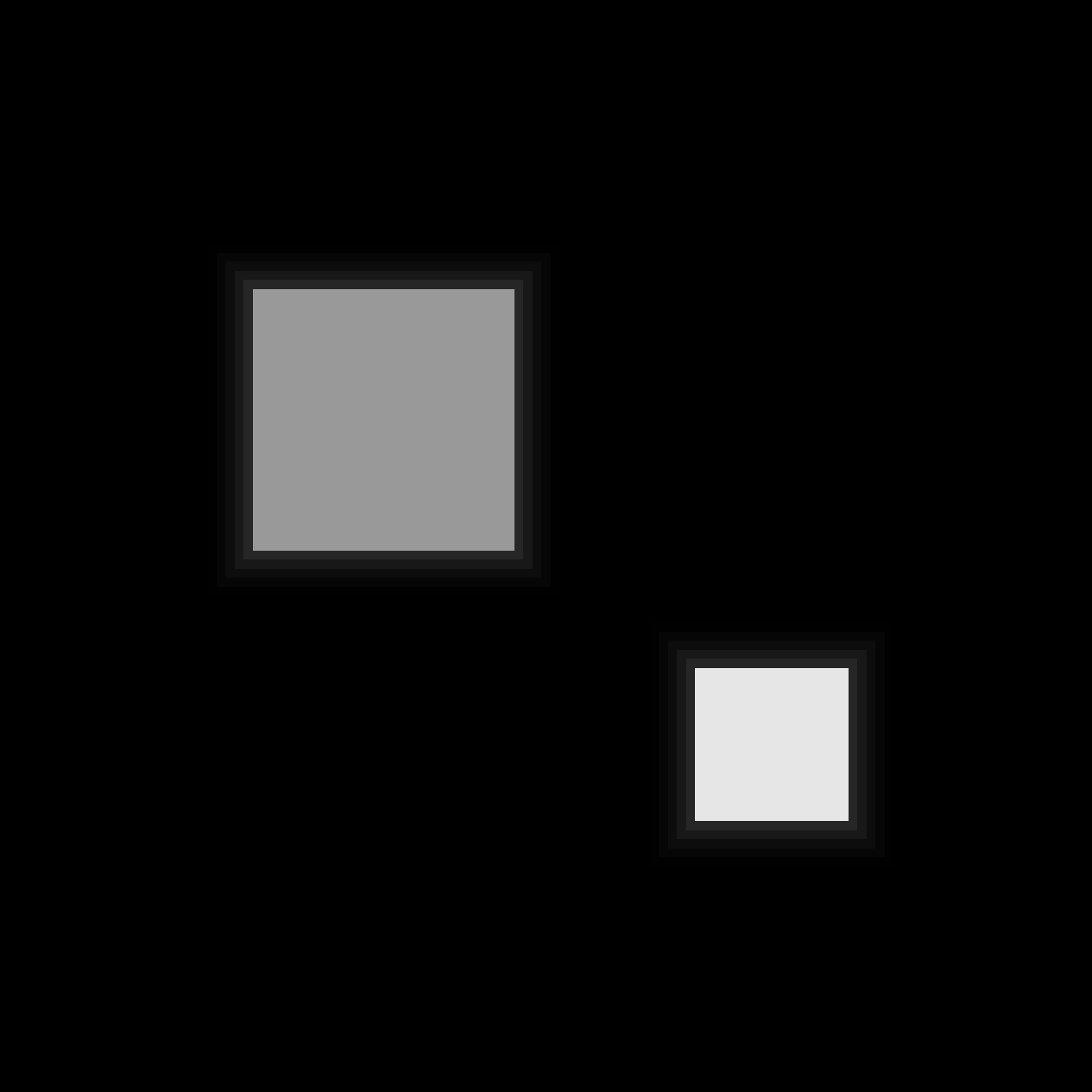}
    \label{fig:toy_erosion}%
  }
  \caption{\textbf{Illustration of classical non-flat grayscale morphology.} A synthetic $121\times121$ grayscale image contains two square regions with intensities $0.60$ and $0.90$ on a zero-valued background. Morphological dilation and erosion are applied using an $11\times11$ spatial support $S_k$ together with a quadratic non-flat structuring function $b(q)$ centered at $q=(0,0)$. The structuring function has its maximum value at the center and decreases with squared distance from the center, reaching $-0.30$ at the corners of the support. Since both square regions are brighter than the surrounding background, non-flat dilation expands their spatial extent, whereas non-flat erosion contracts them.}
  \label{fig:morphology_toy_example}
  \end{center}
\end{figure}

In that regard, dilation and erosion are two fundamental operations 
of mathematical morphology~\cite{serra2023mathematical}. Let $I$ 
denote a 3D grayscale image, $S_k$ the spatial support of a 
$k \times k \times k$ structuring element, and $b(q)$ the value 
of a non-flat structuring function at offset $q$. Classical non-flat grayscale dilation and erosion are respectively defined as
\begin{equation}
\delta_b(I)(v)
=
\max_{q \in S_k}
\left[I(v-q)+b(q)\right],
\label{eq:dilation}
\end{equation}
and
\begin{equation}
\varepsilon_b(I)(v)
=
\min_{q \in S_k}
\left[I(v+q)-b(q)\right],
\label{eq:erosion}
\end{equation}
\noindent where $v$ denotes a voxel location and $q$ indexes the 
spatial offsets within $S_k$. The effect of these operations is 
illustrated in Fig.~\ref{fig:morphology_toy_example} using a 2D synthetic example.

\begin{figure}[ht]
    \begin{center}
   \includegraphics[width=0.8\linewidth]{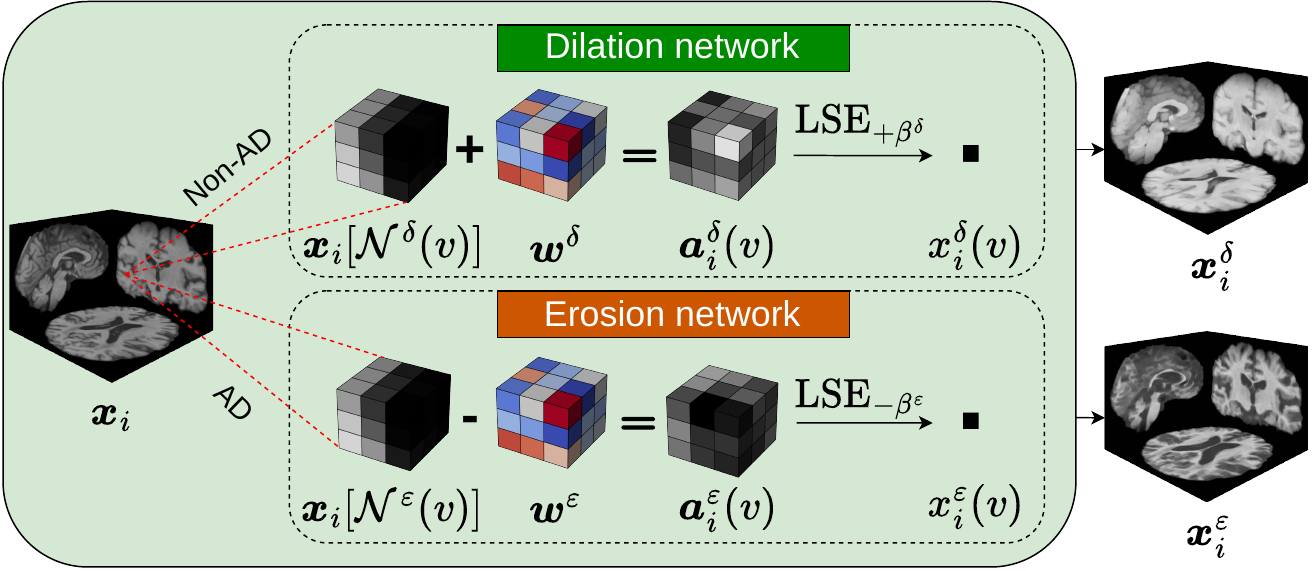}
      
\caption{\textbf{Overview of per-voxel morphological processing.} For an input 3D sMRI $\boldsymbol{x}_i$, local $3\times3\times3$ neighborhoods are extracted around each voxel and routed to the class-conditional pseudo-dilation or pseudo-erosion branch. In the pseudo-dilation branch, used for non-AD samples, the local patch is added to the learnable structuring weights $\boldsymbol{w}^{\delta}$ and reduced using $\operatorname{LSE}_{+\beta^\delta}$ to approximate a smooth maximum. In the pseudo-erosion branch, used for AD samples, the local patch is combined with $\boldsymbol{w}^{\varepsilon}$ by subtraction and reduced using $\operatorname{LSE}_{-\beta^\varepsilon}$ to approximate a smooth minimum. The parameters $\beta^\delta$ and $\beta^\varepsilon$ control the sharpness of the two approximations. Applying these operations over all voxel locations produces the pseudo-dilated volume $\boldsymbol{x}_i^{\delta}$ or pseudo-eroded volume $\boldsymbol{x}_i^{\varepsilon}$.}

        \label{fig:training_pipeline_morpohlogy}
    \end{center}
\end{figure}

In more detail, for a training batch of 3D sMRI images $\{\boldsymbol{x}_i\}_{i=1}^{B}$, where $\boldsymbol{x}_i \in \mathbb{R}^{C \times D \times H \times W}$, each sample is routed to the morphological branch associated with its class. Specifically, non-AD samples are processed by the pseudo-dilation module $\delta$, which introduces the effect of local expansion, to generate the pseudo-dilated volume $\boldsymbol{x}_i^{\delta}$, whereas AD samples are processed by the pseudo-erosion module $\varepsilon$, which produces local morphological contraction, to generate the pseudo-eroded volume ${\boldsymbol{x}}_i^{\varepsilon}$, as shown in Fig.~\ref{fig:training_pipeline}. This design increases intra-class morphological variability while preserving the intended class-specific direction of the augmentation.

As visualized in Fig.~\ref{fig:training_pipeline_morpohlogy}, for each voxel location $v$, a local cubic neighborhood
$\boldsymbol{x}_i[\mathcal{N}^{\delta}(v)]$ or
$\boldsymbol{x}_i[\mathcal{N}^{\varepsilon}(v)]$ of size $k^3$ ($k=3$) is extracted and combined with the learnable structuring weights $\boldsymbol{w}^{\delta}$ or $\boldsymbol{w}^{\varepsilon}$. Since the hard maximum and minimum in Eqs.~\ref{eq:dilation} and~\ref{eq:erosion} are non-differentiable, the resulting local responses are then aggregated using differentiable positive and negative log-sum-exp operators. $\mathrm{LSE}_{+\beta^\delta}$ provides a smooth approximation to the maximum operation in dilation, while $\mathrm{LSE}_{-\beta^\varepsilon}$ provides a smooth approximation to the minimum operation in erosion. The branch-specific sharpness parameters $\beta^\delta$ and $\beta^\varepsilon$ 
control how closely the pseudo-dilation and pseudo-erosion operations approach 
classical grayscale morphology, while remaining trainable end-to-end.

\begin{equation}
\operatorname{LSE}_{+\beta^\delta}(\boldsymbol{s})
=
\frac{1}{\beta^\delta}
\log
\sum_{n=1}^{k^3}
e^{\beta^\delta s_n},
\end{equation}
and
\begin{equation}
\operatorname{LSE}_{-\beta^\varepsilon}(\boldsymbol{s})
=
-\frac{1}{\beta^\varepsilon}
\log
\sum_{n=1}^{k^3}
e^{-\beta^\varepsilon s_n}.
\end{equation}

\noindent Here, $s_n$ denotes the elements of the local response vector, corresponding to
$x_i(v-q)+w^{\delta}(q)$ for pseudo-dilation and
$x_i(v+q)-w^{\varepsilon}(q)$ for pseudo-erosion. Although $\operatorname{LSE}{+\beta^\delta}$ approaches the hard maximum and
$\operatorname{LSE}{-\beta^\varepsilon}$ approaches the hard minimum as
$\beta^\delta, \beta^\varepsilon \rightarrow \infty$, in practice both parameters remain finite throughout training. Consequently, the resulting operations are smooth, differentiable approximations of classical hard dilation and erosion rather than exact equivalents.

Unlike the handcrafted structuring functions in Eqs.~\ref{eq:dilation} and~\ref{eq:erosion}, the proposed modules assign a learnable weight to each offset within the local morphological kernel, enabling the modules to learn the non-flat structuring weights directly from data. This is particularly relevant for 3D brain sMRI, where structural morphology may vary across patients, brain regions, and acquisition domains, motivating the use of learnable rather than fixed structuring weights. The learned structuring weights are shared across spatial locations; therefore, the transformation is class-specific and locally image-dependent, but does not explicitly learn region-specific morphological kernels.

\begin{figure}[ht]
  \begin{center}
  \subfloat[Original non-AD sMRI]{%
    \includegraphics[width=0.32\linewidth]{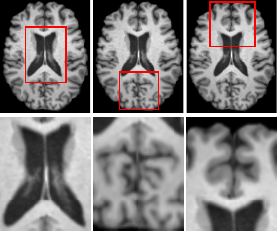}
    \label{fig:dilation_original}%
  }
  \subfloat[Pseudo-dilated non-AD sMRI]{%
    \includegraphics[width=0.32\linewidth]{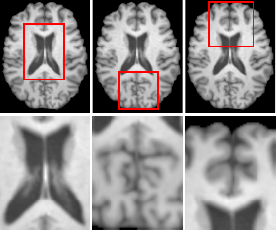}
    \label{fig:dilation_transformed}%
  }

  \vspace{0.5em}

  \subfloat[Original AD sMRI]{%
    \includegraphics[width=0.32\linewidth]{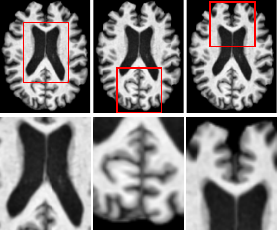}
    \label{fig:erosion_original}%
  }
  \subfloat[Pseudo-eroded AD sMRI]{%
    \includegraphics[width=0.32\linewidth]{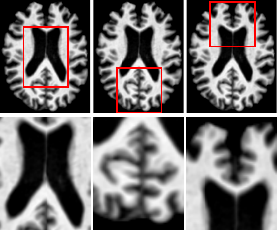}
    \label{fig:erosion_transformed}%
  }

  \caption{\textbf{Effects of learnable class-conditioned morphological augmentation on sMRI slices.} 
  Panels (a) and (b) show an original non-AD sample and its pseudo-dilated counterpart, while panels (c) and (d) show an original AD sample and its pseudo-eroded counterpart. 
  Red boxes and zoomed-in views highlight local structural changes, including expansion and contraction of local boundaries after pseudo-dilation and pseudo-erosion respectively.}
  \label{fig:morph_visualization}
   \end{center}
\end{figure}

In the proposed method, for each non-AD sample $\boldsymbol{x}_i$, the pseudo-dilation module forms the local activation vector

\begin{equation}
\boldsymbol{a}_i^{\delta}(v)
=
\boldsymbol{x}_i[\mathcal{N}^{\delta}(v)]
+
\boldsymbol{w}^{\delta},
\end{equation}
whereas, for each AD sample $\boldsymbol{x}_i$, the pseudo-erosion module forms
\begin{equation}
\boldsymbol{a}_i^{\varepsilon}(v)
=
\boldsymbol{x}_i[\mathcal{N}^{\varepsilon}(v)]
-
\boldsymbol{w}^{\varepsilon},
\end{equation}
\noindent where $\mathcal{N}^{\delta}(v)$ and $\mathcal{N}^{\varepsilon}(v)$ denote the corresponding local neighborhoods centered at voxel $v$, sampled as $x_i(v-q)$ for pseudo-dilation and $x_i(v+q)$ for pseudo-erosion, respectively.

The smooth approximation to the pseudo-dilated voxel value is then computed as 

\begin{equation}
\begin{aligned}
x_i^{\delta}(v)
&=
\frac{1}{\beta^\delta}
\log
\sum_{q \in S_k}
e^{\beta^\delta\left[x_i(v-q)+w^{\delta}(q)\right]} \\
&=
\operatorname{LSE}_{+\beta^\delta}
\left(\boldsymbol{a}_i^{\delta}(v)\right).
\end{aligned}
\end{equation}

\noindent Similarly, the smooth approximation to the pseudo-eroded voxel value is computed as
\begin{equation}
\begin{aligned}
x_i^{\varepsilon}(v)
&=
-\frac{1}{\beta^\varepsilon}
\log
\sum_{q \in S_k}
e^{-\beta^\varepsilon\left[x_i(v+q)-w^{\varepsilon}(q)\right]} \\
&=
\operatorname{LSE}_{-\beta^\varepsilon}
\left(\boldsymbol{a}_i^{\varepsilon}(v)\right).
\end{aligned}
\end{equation}

\noindent Here, $w^{\delta}(q)$ and $w^{\varepsilon}(q)$ denote the learnable weights corresponding to offset $q$ in the pseudo-dilation and pseudo-erosion structuring elements, respectively. To ensure that $\beta^{\delta}$ and $\beta^{\varepsilon}$ remain strictly positive during training, both are initialized to 25 and parameterized using a softplus transformation, thereby guaranteeing valid $\operatorname{LSE}$ behavior throughout optimization.

Applying these operations voxel-wise over the entire sMRI volume produces the learned pseudo-dilated image $\boldsymbol{x}_i^{\delta}$ and the learned pseudo-eroded image $\boldsymbol{x}_i^{\varepsilon}$. As illustrated in Fig.~\ref{fig:morph_visualization}, pseudo-dilation expands local structural boundaries and redistributes neighboring intensities, whereas pseudo-erosion contracts local boundaries and produces an inward redistribution of high-intensity structures.

The original and class-conditioned morphological views are then jointly used with supervised contrastive learning to encourage class-discriminative representations. For each input $\boldsymbol{x}_i$ and its corresponding morphological view $\boldsymbol{x}_i^{m}$, where $m=\delta$ for non-AD samples and $m=\varepsilon$ for AD samples, the encoder $f_\theta$ produces representations $\boldsymbol{z}_i$ and $\boldsymbol{z}_i^{m}$, respectively. These representations are projected by $h_\gamma$ and $\ell_2$-normalized. For a batch containing $B$ original samples and their $B$ corresponding morphological views, the resulting $2B$ normalized embeddings are denoted by $\{\boldsymbol{q}_i\}_{i=1}^{2B}$, with corresponding class labels $y_i$. The weighted supervised contrastive loss is defined as
\begin{equation}
\mathcal{L}_{\mathrm{SupCon}}
=
-\frac{1}{2B}
\sum_{i=1}^{2B}
\frac{\omega_{y_i}}{|P(i)|}
\sum_{j \in P(i)}
\log
\frac{
\exp(\boldsymbol{q}_i^\top \boldsymbol{q}_j/\tau)
}{
\sum_{\substack{k=1 \\ k\neq i}}^{2B}
\exp(\boldsymbol{q}_i^\top \boldsymbol{q}_k/\tau)
},
\end{equation}

\noindent where $P(i)$ is the set of indices of positive samples sharing the same class label as anchor sample $i$, $\tau>0$ is the temperature parameter, and $\omega_{y_i}$ is the class weight used to account for class imbalance, with $\omega_0$ and $\omega_1$ corresponding to the non-AD and AD classes, respectively.

Alongside the contrastive objective, the original encoder representation $\boldsymbol{z}_i$ is passed to the classifier $g_\phi$ to compute the weighted binary cross-entropy loss $\mathcal{L}_{\mathrm{BCE}}$. The overall training objective is
\begin{equation}
\mathcal{L}_{\mathrm{Total}}=\mathcal{L}_{\mathrm{BCE}}+\lambda\mathcal{L}_{\mathrm{SupCon}},
\end{equation}
\noindent where $\lambda$ controls the relative contribution of the supervised contrastive objective. We set $\lambda=0.5$ based exclusively on performance on the NACC validation split, without using any target-domain data for hyperparameter selection. A post-hoc sensitivity analysis examining the effect of different $\lambda$ values across the external cohorts is reported in Table~\ref{tab:hyperparam_sensitivity} of Section~\ref{sec:ablation}.

\section{Experiments}
To assess the effectiveness of the proposed approach, experiments have been conducted against various baseline methods. The goal has been to evaluate model performance in AD classification and assess its generalization across cohorts.

\subsection{Datasets}
Four publicly available cohorts were utilized in this study: the National Alzheimer’s Coordinating Center (NACC)~\cite{NACC}, the Alzheimer’s Disease Neuroimaging Initiative (ADNI)~\cite{ADNI}, the Australian Imaging, Biomarkers, and Lifestyle (AIBL) Study~\cite{AIBL}, and the Open Access Series of Imaging Studies (OASIS-3)~\cite{OASIS}. Each cohort includes 3D sMRI scans with diagnostic labels for normal cognition (NC), mild cognitive impairment (MCI), and dementia due to AD. For binary classification, participants with dementia due to AD constituted the AD group, while NC and MCI participants were grouped as non-AD. Here, non-AD denotes the absence of a clinical diagnosis of AD dementia at the selected visit and does not imply the absence of underlying AD-related pathology.

For each cohort, only T1-weighted MRI scans acquired at 1.5 or 3 Tesla were included. Eligible participants were aged 55 years or older, had a definitive diagnosis, and had an MRI acquired within $\pm$6 months of the corresponding clinical diagnosis. One eligible 3D sMRI scan was retained per participant to prevent subject overlap across data partitions. All imaging and clinical data were obtained in de-identified form under the applicable data-use agreements, and no new participant data were collected. Demographic and class distributions are summarized in Table~\ref{table_demographics_binary}. All sMRI volumes were preprocessed using a standardized pipeline adapted from~\cite{qiu2022multimodal}. Scans were first reoriented to the MNI template space, followed by brain extraction using FSL BET~\cite{smith2002fast}. A two-step linear registration procedure was then applied, including affine registration to MNI-152 space and iterative refinement of skull stripping and alignment. Finally, N4 bias field correction was used to reduce intensity inhomogeneity across subjects.

\begin{table}[ht]
\caption{\textbf{Participant demographics across sMRI cohorts.} Data are reported for four independent cohorts: NACC, ADNI, AIBL, and OASIS-3. Class distribution, mean age, and male participant counts are reported for the two diagnostic categories (Non-AD and AD) where available.}

\label{table_demographics_binary}
\begin{center}
\setlength{\tabcolsep}{6pt} 
\scriptsize
\begin{tabular}{l c c c}
\hline
\textbf{Dataset} & \textbf{Group (Participants)} & \textbf{Age, years} & \textbf{Gender} \\
                 &                               & \textbf{mean $\pm$ std} & \textbf{male count} \\
\hline
\multirow{2}{*}{NACC \cite{NACC}} 
& Non-AD (n=3699) & 71.1 $\pm$ 9.7 & 1426 (38.6\%) \\
& AD (n=948)      & 75.0 $\pm$ 9.1 & 431 (45.5\%) \\
\hline
\multirow{2}{*}{ADNI \cite{ADNI}} 
& Non-AD (n=1452) & 73.3 $\pm$ 7.3 & 807 (55.6\%) \\
& AD (n=369)      & 74.9 $\pm$ 7.8 & 203 (55.0\%) \\
\hline
\multirow{2}{*}{AIBL \cite{AIBL}} 
& Non-AD (n=582)  & 72.9 $\pm$ 6.5 & 256 (44.0\%) \\
& AD (n=79)       & 73.3 $\pm$ 7.8 & 33 (41.8\%) \\
\hline
\multirow{2}{*}{OASIS-3 \cite{OASIS}} 
& Non-AD (n=451)  & NA & NA \\
& AD (n=193)      & NA & NA \\
\hline
\end{tabular}
\end{center}
\end{table}

Despite uniform preprocessing, substantial variability remained across cohorts due to differences in imaging hardware, acquisition protocols, cohort characteristics, and the relative NC/MCI composition of the non-AD class. This is reflected in the t-SNE visualizations in Fig.~\ref{fig:tsne_combined} (Section~\ref{sec:ablation}), where 3D U-Net features form distinct dataset-specific clusters, highlighting the challenge of the SDG setting.

\begin{table}[t]
\caption{\textbf{Cross-dataset results on external sMRI cohorts.} Models were trained on NACC and evaluated on OASIS-3, ADNI, and AIBL using the seed-42 run. Best results are in bold.}
\label{tab:cross_dataset}
\centering
\setlength{\tabcolsep}{2pt}
\scriptsize
\resizebox{\textwidth}{!}{%
\begin{tabular}{l|ccccc|ccccc|ccccc}
\hline
\textbf{Method}
& \multicolumn{5}{c|}{\textbf{OASIS-3}}
& \multicolumn{5}{c|}{\textbf{ADNI}}
& \multicolumn{5}{c}{\textbf{AIBL}} \\
\hline
& \textbf{ACC} & \textbf{F1} & \textbf{SEN} & \textbf{SPE} & \textbf{MCC}
& \textbf{ACC} & \textbf{F1} & \textbf{SEN} & \textbf{SPE} & \textbf{MCC}
& \textbf{ACC} & \textbf{F1} & \textbf{SEN} & \textbf{SPE} & \textbf{MCC} \\
\hline
Baseline \cite{Models_Genesis}
& 70.96 & 0.697 & \textbf{0.850} & 0.650 & 0.458
& 64.47 & 0.586 & 0.664 & 0.640 & 0.247
& 79.43 & 0.665 & 0.722 & 0.804 & 0.389 \\
CCSDG \cite{CCSDG}
& 72.20 & 0.700 & 0.756 & 0.707 & 0.429
& 66.34 & 0.588 & 0.583 & 0.684 & 0.222
& 72.31 & 0.599 & 0.696 & 0.727 & 0.293 \\
Mixup \cite{Mixup}
& 73.76 & 0.715 & 0.762 & 0.727 & 0.454
& 68.37 & 0.613 & 0.634 & 0.696 & 0.275
& 79.27 & 0.652 & 0.658 & 0.811 & 0.354 \\
RSC \cite{paper11}
& 72.05 & 0.704 & 0.813 & 0.681 & 0.454
& 65.84 & 0.595 & 0.648 & 0.661 & 0.253
& 78.52 & 0.654 & 0.709 & 0.796 & 0.371 \\
EFDM \cite{zhang2022exact}
& 74.84 & 0.724 & 0.756 & 0.745 & 0.468
& 63.32 & 0.568 & 0.602 & 0.641 & 0.199
& 84.27 & 0.694 & 0.608 & 0.875 & 0.404 \\
Mixstyle \cite{zhou2021mixstyle}
& 72.36 & 0.710 & 0.845 & 0.672 & 0.473
& 67.00 & 0.607 & 0.667 & 0.671 & 0.277
& 71.56 & 0.600 & 0.747 & 0.711 & 0.313 \\
EM \cite{batool2026higher}
& 69.41 & 0.681 & 0.819 & 0.641 & 0.421
& 60.08 & 0.552 & 0.667 & 0.584 & 0.202
& 69.89 & 0.587 & 0.747 & 0.692 & 0.297 \\
AAGN \cite{jiang2024anatomy}
& 73.60 & 0.720 & 0.834 & 0.694 & 0.485
& 63.48 & 0.581 & \textbf{0.680} & 0.623 & 0.246
& 69.44 & 0.592 & \textbf{0.810} & 0.679 & 0.327 \\
FM-SDG \cite{li2023frequency}
& 78.11 & 0.751 & 0.720 & 0.807 & 0.506
& 78.97 & 0.701 & 0.604 & 0.837 & 0.408
& 77.61 & 0.633 & 0.633 & 0.796 & 0.319 \\
DeTTA \cite{wen2024denoising}
& 75.78 & 0.731 & 0.741 & 0.765 & 0.476
& 78.58 & 0.701 & 0.623 & 0.827 & 0.411
& 78.52 & 0.639 & 0.620 & 0.808 & 0.323 \\
ViT-LD \cite{kunanbayev2024training}
& 71.58 & 0.687 & 0.689 & 0.727 & 0.390
& 78.25 & 0.696 & 0.612 & 0.826 & 0.400
& 74.13 & 0.607 & 0.658 & 0.753 & 0.292 \\
ADA-SGD \cite{zhang2025adversarial}
& 69.72 & 0.673 & 0.710 & 0.692 & 0.372
& 68.75 & 0.619 & 0.650 & 0.697 & 0.289
& 73.07 & 0.599 & 0.658 & 0.741 & 0.280 \\
DCON \cite{wang2026hybrid}
& 65.99 & 0.633 & 0.653 & 0.663 & 0.292
& 67.11 & 0.601 & 0.623 & 0.683 & 0.254
& 77.61 & 0.630 & 0.620 & 0.797 & 0.312 \\
\textbf{Ours}
& \textbf{79.35} & \textbf{0.765} & 0.741 & \textbf{0.816} & \textbf{0.535}
& \textbf{80.18} & \textbf{0.721} & 0.650 & \textbf{0.840} & \textbf{0.450}
& \textbf{87.44} & \textbf{0.748} & 0.696 & \textbf{0.899} & \textbf{0.511} \\
\hline
\end{tabular}%
}
\end{table}

\subsection{Experimental Settings}
All experiments were conducted on an NVIDIA A6000 GPU with 48 GB of memory. Due to the memory requirements of 3D sMRI processing, a batch size of 2 was used with gradient accumulation to obtain an effective batch size of 16. Input volumes had dimensions of $182 \times 218 \times 182$ voxels. The model was optimized using stochastic gradient descent with momentum 0.9. The encoder and projector used a learning rate of $1\times10^{-2}$ and weight decay of $5\times10^{-4}$, while the pseudo-dilation and pseudo-erosion kernels used a learning rate of $1\times10^{-3}$ and weight decay of $1\times10^{-5}$. The morphological kernels had size $3\times3\times3$, and the learnable sharpness parameters $\beta^\delta$ and $\beta^\varepsilon$ were optimized with a learning rate of $1\times10^{-2}$ and no weight decay. The supervised contrastive loss used a temperature parameter $\tau=0.07$. A multi-step scheduler reduced the learning rate by a factor of 0.1 at epochs 25, 45, and 58, and training was performed for 60 epochs. Class imbalance was addressed using weighted binary cross-entropy, with the AD class weighted according to the inverse class-frequency ratio.

To evaluate single domain generalization, the model was trained and validated exclusively on NACC using a stratified 80/20 split and evaluated on three unseen target datasets, ADNI, AIBL, and OASIS-3, following the standard SDG protocol \cite{qiao2020learning}. The proposed approach was compared with a 3D U-Net baseline \cite{Models_Genesis} and several DG methods. Hyperparameters for competing methods followed their original implementations where applicable. MixUp \cite{Mixup} used $\alpha=0.3$; RSC \cite{paper11} used 20\% feature dropout, 5\% background dropout, and probability 0.3; MixStyle \cite{zhou2021mixstyle} used $\alpha=0.1$ and probability 0.5; EM \cite{batool2026higher} used $\alpha=0.7$ and probability 0.9; EFDM \cite{zhang2022exact} used patch replacement probability $p=0.5$ and interpolation factor $\alpha=0.1$; and CCSDG with Feature Distribution Alignment \cite{CCSDG} used a low-frequency spectrum replacement ratio $L\sim[0.05,0.1]$. AAGN \cite{jiang2024anatomy}, FM-SDG \cite{li2023frequency}, DeTTA \cite{wen2024denoising}, ViT-LD \cite{kunanbayev2024training}, ADA-SGD \cite{zhang2025adversarial}, and DCON \cite{wang2026hybrid} followed their original configurations where applicable, with task-specific adaptations as needed. For the multi-seed and post-hoc statistical analyses, AAGN \cite{jiang2024anatomy} was selected for its higher sensitivity on ADNI and AIBL, while FM-SDG \cite{li2023frequency} was selected as the strongest competing method based on cohort-averaged F1. Performance was evaluated using accuracy, macro-F1-score, sensitivity, specificity, and Matthews correlation coefficient (MCC).

\subsection{Results and Discussion}
Table~\ref{tab:cross_dataset} summarizes cross-dataset generalization from NACC to the three external cohorts. Overall, the proposed method achieves the highest numerical F1-score, specificity, MCC, and accuracy across all three target datasets. When averaged across the external cohorts, its F1-score exceeds that of the strongest competing method by 5.0 percentage points. These results indicate that the proposed framework provides a favorable balance between the AD and non-AD classes under domain shift. Although some competing methods achieve higher sensitivity, this is generally accompanied by lower specificity and weaker overall discrimination, suggesting a different sensitivity-specificity trade-off.

\begin{table}[t]
\caption{\textbf{Five-seed F1 summaries and pairwise statistical comparisons across external cohorts.}
Mean $\pm$ standard deviation (SD) summarizes F1 across five random seeds.
Pairwise $\Delta$F1, 95\% confidence intervals (CIs), and $p$-values are from paired stratified bootstrap analysis with 10,000 resamples.
Holm correction was applied across all nine pairwise comparisons.
Values in bold indicate statistical significance after Holm correction ($p_{\mathrm{Holm}}<0.05$).}
\label{tab:bootstrap_significance}
\centering
\setlength{\tabcolsep}{4pt}
\scriptsize

\begin{tabular}{llcccccc}
\hline
\textbf{Dataset} &
\textbf{Comparison} &
\textbf{Method 1 F1} &
\textbf{Method 2 F1} &
$\boldsymbol{\Delta}$\textbf{F1} &
\textbf{95\% CI} &
\textbf{Raw $p$} &
\textbf{Holm $p$} \\
& & \textbf{(mean $\pm$ SD)} &
\textbf{(mean $\pm$ SD)} & & & & \\
\hline

\multirow{3}{*}{ADNI}
& Ours vs Baseline
& $0.699 \pm 0.014$ & $0.611 \pm 0.042$ & 0.088
& [0.070, 0.106] & 0.001 & \textbf{0.009} \\
& Ours vs AAGN
& $0.699 \pm 0.014$ & $0.614 \pm 0.019$ & 0.085
& [0.064, 0.104] & 0.001 & \textbf{0.009} \\
& Ours vs FM-SDG
& $0.699 \pm 0.014$ & $0.682 \pm 0.019$ & 0.017
& [-0.003, 0.038] & 0.106 & 0.515 \\
\hline

\multirow{3}{*}{AIBL}
& Ours vs Baseline
& $0.674 \pm 0.042$ & $0.643 \pm 0.013$ & 0.031
& [-0.005, 0.072] & 0.106 & 0.515 \\
& Ours vs AAGN
& $0.674 \pm 0.042$ & $0.600 \pm 0.005$ & 0.074
& [0.022, 0.125] & 0.009 & 0.054 \\
& Ours vs FM-SDG
& $0.674 \pm 0.042$ & $0.628 \pm 0.007$ & 0.046
& [-0.009, 0.100] & 0.103 & 0.515 \\
\hline

\multirow{3}{*}{OASIS-3}
& Ours vs Baseline
& $0.717 \pm 0.027$ & $0.677 \pm 0.013$ & 0.040
& [0.010, 0.070] & 0.005 & \textbf{0.035} \\
& Ours vs AAGN
& $0.717 \pm 0.027$ & $0.703 \pm 0.014$ & 0.014
& [-0.028, 0.061] & 0.511 & 1.000 \\
& Ours vs FM-SDG
& $0.717 \pm 0.027$ & $0.714 \pm 0.022$ & 0.003
& [-0.039, 0.050] & 0.860 & 1.000 \\
\hline

\end{tabular}
\end{table}

This pattern is also reflected across individual cohorts. The proposed method performs consistently on ADNI, while strong overall results are also observed on OASIS-3 and AIBL. The lower sensitivity observed relative to some competing methods may arise from differences between the source and target cohorts in acquisition protocols, class composition, and the relative proportions of NC and MCI participants within the non-AD group. Such differences can shift the decision boundary learned from NACC and affect sensitivity and specificity differently across external populations. The strong MCC and F1 results nevertheless indicate that the proposed model maintains balanced discrimination despite this heterogeneity.

\zb{To assess training stability, the baseline \cite{Models_Genesis}, AAGN \cite{jiang2024anatomy}, FM-SDG \cite{li2023frequency}, and the proposed method were evaluated over five random seeds (42, 1, 2, 3, and 4); their mean F1 scores and standard deviations, together with paired-bootstrap comparisons, are reported in Table~\ref{tab:bootstrap_significance}. Because of the computational cost of 3D sMRI training, five-seed evaluations were limited to the statistical analysis; all other experiments report results from the seed-42 run. Pairwise statistical comparisons were conducted using paired stratified bootstrap resampling at the patient level, with 10,000 bootstrap resamples drawn with replacement \cite{efron1992bootstrap}. The same resampled patients were used across methods, and F1 was averaged across the five random seeds before pairwise differences were computed. Two-sided 95\% confidence intervals and $p$-values were reported. Two-sided $p$-values were estimated as twice the smaller of the two empirical bootstrap tail probabilities. Holm correction was applied across all nine pairwise comparisons \cite{holm1979simple}. Statistical significance was defined at $p \leq 0.05$. The repeated-run analysis demonstrates the effectiveness of the proposed framework, with statistically significant F1 improvements over the baseline on ADNI ($p = 0.009$) and OASIS-3 (p = 0.035), and over AAGN on ADNI (p = 0.009), after Holm correction. The proposed method also achieves the highest numerical mean F1 on AIBL and consistently higher mean F1 than FM-SDG across all three cohorts. Although these additional improvements do not reach statistical significance, they indicate favorable numerical performance against strong competing methods. On AIBL, the relatively small number of AD participants (79, compared with 369 in ADNI) may increase sensitivity to individual prediction changes and contribute to seed-to-seed variability, although this remains a possible explanation. Overall, the results support the effectiveness of the proposed approach while indicating that the magnitude and statistical support of improvements vary across cohorts.}

\subsection{Ablation Study}
\label{sec:ablation}

Table~\ref{tab:ablation_combined} evaluates the contribution of the main components across OASIS-3, ADNI, and AIBL. Neither SupCon nor morphology alone provides consistent gains over the baseline across all cohorts, whereas their combination in the proposed framework yields substantially stronger and more consistent performance. Replacing the learnable morphology with fixed operations, removing the soft approximation, or removing class-conditional processing generally degrades performance. Furthermore, reversing the class-specific morphological operations leads to considerable performance reductions, highlighting the importance of assigning appropriate operations to each diagnostic class. Randomly assigning morphological operations also results in lower performance than the proposed configuration, further supporting the role of class-conditional processing. Overall, the proposed framework achieves the highest numerical ACC, F1, SPE, and MCC across all three cohorts, suggesting complementary contributions from learnable class-conditioned morphology and supervised contrastive learning.
\begin{table}[t]
\caption{\textbf{Ablation study across external sMRI cohorts.} Best results are in bold.}
\label{tab:ablation_combined}
\centering
\setlength{\tabcolsep}{2pt}
\scriptsize
\resizebox{\textwidth}{!}{%
\begin{tabular}{l|ccccc|ccccc|ccccc}
\hline
\textbf{Configuration}
& \multicolumn{5}{c|}{\textbf{OASIS-3}}
& \multicolumn{5}{c|}{\textbf{ADNI}}
& \multicolumn{5}{c}{\textbf{AIBL}} \\
\hline
& \textbf{ACC} & \textbf{F1} & \textbf{SEN} & \textbf{SPE} & \textbf{MCC}
& \textbf{ACC} & \textbf{F1} & \textbf{SEN} & \textbf{SPE} & \textbf{MCC}
& \textbf{ACC} & \textbf{F1} & \textbf{SEN} & \textbf{SPE} & \textbf{MCC} \\
\hline
Baseline \cite{Models_Genesis}
& 70.96 & 0.697 & \textbf{0.850} & 0.650 & 0.458
& 64.47 & 0.586 & \textbf{0.664} & 0.640 & 0.247
& 79.43 & 0.665 & 0.722 & 0.804 & 0.389 \\
Baseline + SupCon
& 65.37 & 0.634 & 0.699 & 0.634 & 0.306
& 62.60 & 0.569 & 0.645 & 0.621 & 0.216
& 80.94 & 0.679 & 0.722 & 0.821 & 0.410 \\
Baseline + Morphology
& 66.15 & 0.641 & 0.705 & 0.643 & 0.320
& 62.22 & 0.566 & 0.650 & 0.615 & 0.215
& 80.33 & 0.671 & 0.709 & 0.816 & 0.395 \\
With Fixed Morphology
& 68.01 & 0.660 & 0.731 & 0.659 & 0.358
& 63.76 & 0.581 & \textbf{0.664} & 0.631 & 0.239
& 78.97 & 0.662 & \textbf{0.734} & 0.797 & 0.390 \\
w/o Soft Morphology
& 75.78 & 0.734 & 0.762 & 0.756 & 0.484
& 71.88 & 0.636 & 0.596 & 0.750 & 0.298
& 85.17 & 0.704 & 0.608 & 0.885 & 0.421 \\
w/o Conditional Classes
& 66.93 & 0.649 & 0.720 & 0.647 & 0.338
& 62.99 & 0.569 & 0.629 & 0.630 & 0.211
& 77.46 & 0.642 & 0.696 & 0.785 & 0.351 \\
Reversed Morphology
& 61.96 & 0.600 & 0.668 & 0.599 & 0.245
& 63.15 & 0.564 & 0.588 & 0.643 & 0.189
& 78.67 & 0.653 & 0.696 & 0.799 & 0.366 \\
Random Morphology
& 76.09 & 0.730 & 0.705 & 0.785 & 0.467
& 67.87 & 0.609 & 0.631 & 0.691 & 0.268
& 79.27 & 0.661 & 0.709 & 0.804 & 0.381 \\
\textbf{Ours}
& \textbf{79.35} & \textbf{0.765} & 0.741 & \textbf{0.816} & \textbf{0.535}
& \textbf{80.18} & \textbf{0.721} & 0.650 & \textbf{0.840} & \textbf{0.450}
& \textbf{87.44} & \textbf{0.748} & 0.696 & \textbf{0.899} & \textbf{0.511} \\
\hline
\end{tabular}%
}
\end{table}

Having established the contribution of the individual components, Table~\ref{tab:hyperparam_sensitivity} examines the sensitivity of the framework to the SupCon weight $\lambda$ and the LSE sharpness parameter $\beta$. The value $\lambda=0.5$ was selected exclusively using the NACC validation split, as described in Section~\ref{sec:morphological_augmentation}; the results reported here on the external cohorts are post-hoc and were not used for hyperparameter selection. Smaller or larger values of $\lambda$ produce more dataset-dependent trade-offs, whereas $\lambda=0.5$ maintains a comparatively balanced performance across the three cohorts. Similarly, fixed values of $\beta$ show variable behavior across datasets, while the learnable formulation provides the most consistent overall performance by adapting the sharpness of the morphological operations during training.

\begin{table}[ht]
\caption{\textbf{Post-hoc sensitivity analysis of $\lambda$ and $\beta$ across external sMRI cohorts.} 
Top: effect of the supervised contrastive loss weight $\lambda$. 
Bottom: effect of the $\mathrm{LSE}$ parameter $\beta$ used in the soft morphological operations. 
Best results are in bold. The results indicate that moderate values of $\lambda$ and a learnable $\beta$ provide the most consistent performance across datasets.}
\label{tab:hyperparam_sensitivity}
\begin{center}
\setlength{\tabcolsep}{2.5pt}
\scriptsize
\begin{tabular}{c|ccccc|ccccc|ccccc}
\hline

\multicolumn{16}{c}{\textbf{SupCon Weight ($\lambda$)}} \\
\hline

\multirow{2}{*}{\textbf{$\lambda$}} 
& \multicolumn{5}{c|}{\textbf{OASIS-3}} 
& \multicolumn{5}{c|}{\textbf{ADNI}} 
& \multicolumn{5}{c}{\textbf{AIBL}} \\

& \textbf{ACC} & \textbf{F1} & \textbf{SEN} & \textbf{SPE} & \textbf{MCC} 
& \textbf{ACC} & \textbf{F1} & \textbf{SEN} & \textbf{SPE} & \textbf{MCC} 
& \textbf{ACC} & \textbf{F1} & \textbf{SEN} & \textbf{SPE} & \textbf{MCC} \\
\hline

0.3 
& 74.38 & 0.726 & \textbf{0.813} & 0.714 & 0.486
& 72.60 & 0.665 & 0.737 & 0.723 & 0.383
& 83.21 & 0.698 & \textbf{0.696} & 0.851 & 0.432 \\

0.5 
& \textbf{79.35} & \textbf{0.765} & 0.741 & \textbf{0.816} & \textbf{0.535}
& \textbf{80.18} & \textbf{0.721} & 0.650 & \textbf{0.840} & 0.450
& \textbf{87.44} & \textbf{0.748} & \textbf{0.696} & \textbf{0.899} & \textbf{0.511} \\

0.7 
& 72.20 & 0.679 & 0.596 & 0.776 & 0.361
& 77.10 & 0.710 & \textbf{0.772} & 0.771 & \textbf{0.461}
& 85.48 & 0.720 & 0.671 & 0.880 & 0.459 \\

0.9 
& 77.64 & 0.753 & 0.777 & 0.776 & 0.520
& 75.89 & 0.672 & 0.604 & 0.798 & 0.359
& 82.75 & 0.694 & \textbf{0.696} & 0.845 & 0.425 \\

\hline

\multicolumn{16}{c}{\textbf{LogSumExp ($\beta$)}} \\
\hline

\multirow{2}{*}{\textbf{$\beta$}} 
& \multicolumn{5}{c|}{\textbf{OASIS-3}} 
& \multicolumn{5}{c|}{\textbf{ADNI}} 
& \multicolumn{5}{c}{\textbf{AIBL}} \\

& \textbf{ACC} & \textbf{F1} & \textbf{SEN} & \textbf{SPE} & \textbf{MCC} 
& \textbf{ACC} & \textbf{F1} & \textbf{SEN} & \textbf{SPE} & \textbf{MCC} 
& \textbf{ACC} & \textbf{F1} & \textbf{SEN} & \textbf{SPE} & \textbf{MCC} \\
\hline

10 
& 74.38 & 0.724 & 0.798 & 0.721 & 0.479
& 67.60 & 0.610 & \textbf{0.656} & 0.681 & 0.278
& 82.30 & 0.684 & 0.671 & 0.844 & 0.404 \\

25 
& 78.57 & 0.762 & 0.788 & 0.785 & \textbf{0.539}
& 76.77 & 0.676 & 0.583 & 0.815 & 0.362
& 83.96 & 0.700 & 0.658 & 0.864 & 0.425 \\

40 
& 72.52 & 0.712 & \textbf{0.850} & 0.672 & 0.478
& 71.11 & 0.638 & 0.648 & 0.727 & 0.316
& 77.46 & 0.644 & \textbf{0.709} & 0.784 & 0.358 \\

50 
& 77.17 & 0.749 & 0.782 & 0.767 & 0.514
& 72.43 & 0.643 & 0.607 & 0.754 & 0.311
& 84.87 & 0.717 & 0.696 & 0.869 & 0.460 \\

Learnable 
& \textbf{79.35} & \textbf{0.765} & 0.741 & \textbf{0.816} & 0.535
& \textbf{80.18} & \textbf{0.721} & 0.650 & \textbf{0.840} & \textbf{0.450}
& \textbf{87.44} & \textbf{0.748} & 0.696 & \textbf{0.899} & \textbf{0.511} \\

\hline
\end{tabular}
\end{center}
\end{table}

To further examine whether the observed behavior depends on the choice of source cohort, Table~\ref{table_cross_source_v3} reports cross-source experiments in which each model is trained on a single cohort and evaluated on the remaining unseen cohorts. When ADNI is used as the source, the proposed method numerically improves ACC, F1, SPE, and MCC across all three target cohorts, with the largest accuracy gain of 5.59 percentage points on OASIS-3. With OASIS-3 as the source, numerical improvements in ACC and SPE are observed on AIBL, despite a slight decrease in F1, while ACC, F1, and SPE improve on NACC, including a 4.82 percentage-point accuracy gain. Performance on ADNI remains comparable to the baseline. When AIBL is used as the source, the proposed method achieves higher sensitivity across all three targets, although ACC and F1 are lower than the baseline. This trade-off may partly reflect the smaller and more imbalanced AIBL cohort, which can influence the learned decision boundary. Overall, the numerical results suggest that the benefits of morphology-guided learning depend partly on the source distribution, with the most consistent gains observed when ADNI is used for training.

\begin{table}[ht]
\caption{\textbf{Cross-source domain generalization experiments.} Models are trained on one source cohort and evaluated on the remaining target cohorts. For each source setting, the baseline 3D U-Net is compared with Morph.}
\label{table_cross_source_v3}
\centering
\scriptsize
\setlength{\tabcolsep}{1.5pt}
\resizebox{\textwidth}{!}{%
\begin{tabular}{l|ccccc|ccccc|ccccc|ccccc}
\hline
\textbf{Method} &
\multicolumn{5}{c|}{\textbf{OASIS-3}} &
\multicolumn{5}{c|}{\textbf{ADNI}} &
\multicolumn{5}{c|}{\textbf{AIBL}} &
\multicolumn{5}{c}{\textbf{NACC}} \\
\hline
& \textbf{ACC} & \textbf{F1} & \textbf{SEN} & \textbf{SPE} & \textbf{MCC}
& \textbf{ACC} & \textbf{F1} & \textbf{SEN} & \textbf{SPE} & \textbf{MCC}
& \textbf{ACC} & \textbf{F1} & \textbf{SEN} & \textbf{SPE} & \textbf{MCC}
& \textbf{ACC} & \textbf{F1} & \textbf{SEN} & \textbf{SPE} & \textbf{MCC} \\
\hline
\multicolumn{21}{c}{\textbf{Source Dataset: OASIS-3}} \\
\hline
Baseline \cite{Models_Genesis}
& - & - & - & - & -
& \textbf{68.15} & \textbf{0.595} & 0.542 & \textbf{0.717} & 0.220
& 73.98 & \textbf{0.610} & \textbf{0.684} & 0.747 & \textbf{0.304}
& 73.90 & 0.667 & \textbf{0.669} & 0.757 & 0.365 \\
\textbf{Ours}
& - & - & - & - & -
& 67.66 & 0.593 & \textbf{0.553} & 0.708 & \textbf{0.221}
& \textbf{74.13} & 0.605 & 0.646 & \textbf{0.754} & 0.285
& \textbf{78.72} & \textbf{0.708} & 0.651 & \textbf{0.822} & \textbf{0.427} \\
\hline
\multicolumn{21}{c}{\textbf{Source Dataset: ADNI}} \\
\hline
Baseline \cite{Models_Genesis}
& 61.80 & 0.608 & 0.762 & 0.557 & 0.292
& - & - & - & - & -
& 56.58 & 0.492 & \textbf{0.772} & 0.538 & 0.201
& 63.46 & 0.582 & 0.688 & 0.621 & 0.251 \\
\textbf{Ours}
& \textbf{67.39} & \textbf{0.660} & \textbf{0.788} & \textbf{0.625} & \textbf{0.378}
& - & - & - & - & -
& \textbf{60.97} & \textbf{0.519} & 0.734 & \textbf{0.593} & \textbf{0.213}
& \textbf{65.16} & \textbf{0.596} & \textbf{0.689} & \textbf{0.642} & \textbf{0.270} \\
\hline
\multicolumn{21}{c}{\textbf{Source Dataset: AIBL}} \\
\hline
Baseline \cite{Models_Genesis}
& \textbf{73.91} & \textbf{0.690} & 0.570 & \textbf{0.812} & \textbf{0.380}
& \textbf{73.92} & \textbf{0.602} & 0.377 & \textbf{0.831} & \textbf{0.205}
& - & - & - & - & -
& \textbf{77.79} & \textbf{0.666} & 0.488 & \textbf{0.852} & 0.333 \\
\textbf{Ours}
& 70.96 & 0.675 & \textbf{0.642} & 0.738 & 0.360
& 69.85 & 0.589 & \textbf{0.450} & 0.762 & 0.189
& - & - & - & - & -
& 73.62 & 0.661 & \textbf{0.651} & 0.758 & \textbf{0.352} \\
\hline
\end{tabular}%
}
\end{table}

\begin{figure}[ht]
  \begin{center}
  \subfloat[3D U-Net]{%
    \includegraphics[width=0.47\linewidth]{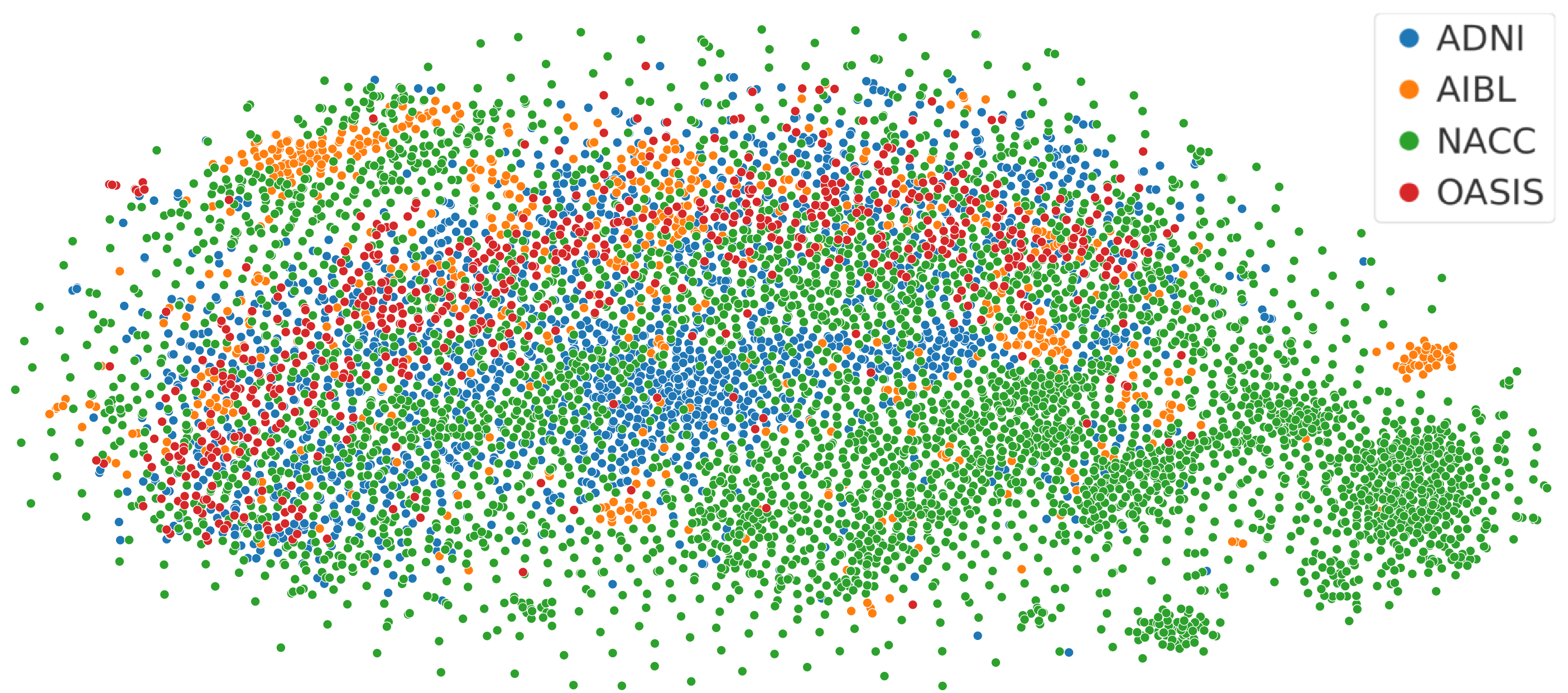}%
    \label{fig:tsne-1}
  }\hfill
  \subfloat[Ours]{%
    \includegraphics[width=0.47\linewidth]{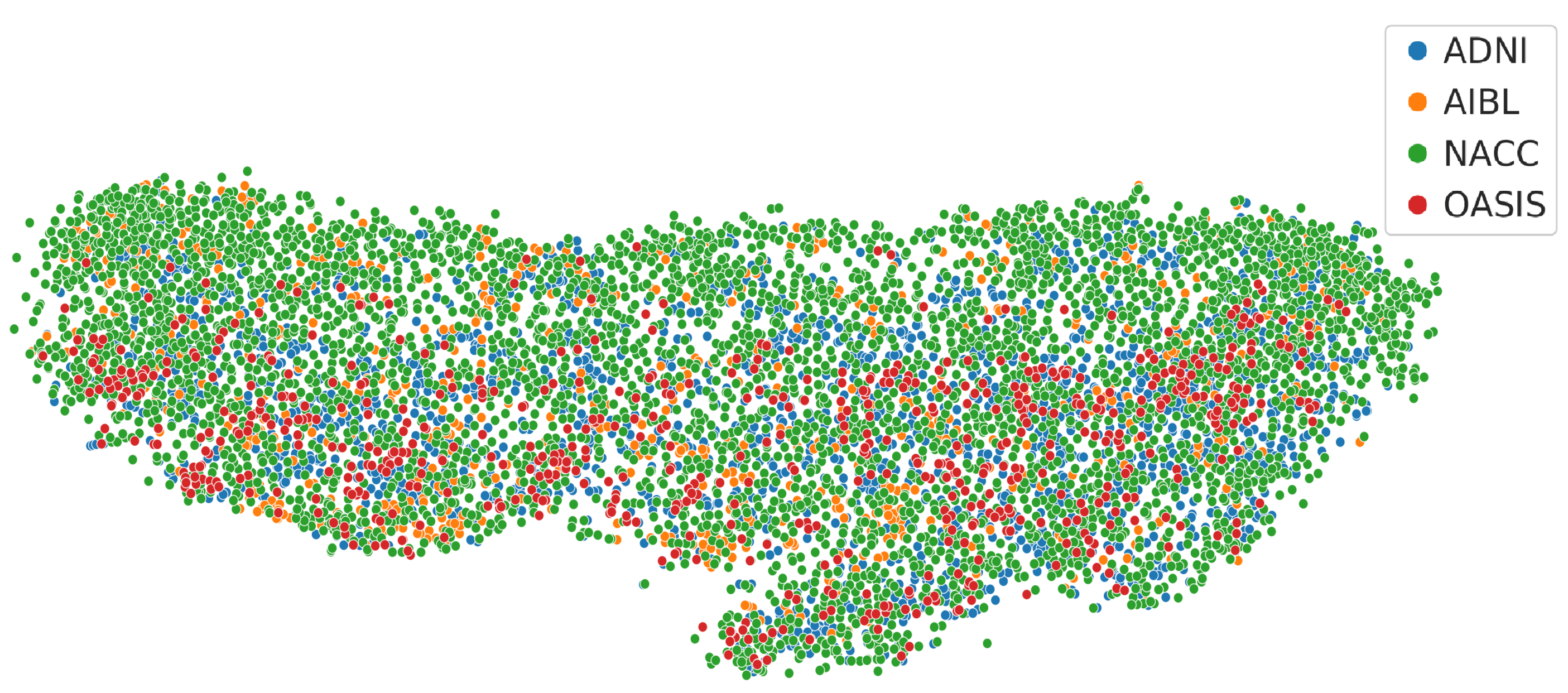}%
    \label{fig:tsne-2}
  }

\caption{\textbf{t-SNE visualizations of sMRI embeddings under different training settings.} Embeddings from NACC, ADNI, AIBL, and OASIS-3 are shown for the baseline 3D U-Net (a) and the proposed Morph model (b). Baseline embeddings were extracted from the final encoder layer, whereas Morph embeddings were taken from the encoder representation before the projection head. The proposed model shows less cohort-specific clustering and a more interleaved embedding structure, suggesting improved cross-dataset alignment.}  \label{fig:tsne_combined}
  \end{center}
\end{figure}

Fig.~\ref{fig:tsne_combined} provides a qualitative view of the learned representations across NACC, ADNI, AIBL, and OASIS-3. The baseline 3D U-Net exhibits more visible cohort-dependent grouping, whereas the proposed model produces a more interleaved embedding structure with less apparent separation between datasets. Although t-SNE is only a qualitative visualization and does not by itself establish domain invariance, the reduced cohort-specific grouping is consistent with the quantitative generalization results and suggests that the proposed training strategy encourages less dataset-dependent representations.

Finally, Table~\ref{tab:computational_complexity} compares the computational requirements of the evaluated models. The proposed method introduces only a modest increase in trainable parameters, while requiring higher training time due to the additional class-conditioned morphological transformations and supervised contrastive learning branch. This added cost is accompanied by improved cross-dataset performance, with the proposed approach achieving higher F1 scores than the baseline and showing numerical improvements over competing methods across the external cohorts. Since the morphological augmentation and projection branches are used only during training, they do not add inference-time computation.

\begin{table}[t]
\caption{\textbf{Computational complexity of the evaluated models.}
Parameter count, computational cost, and training time per epoch are reported for the baseline, AAGN, and the proposed method.}
\label{tab:computational_complexity}
\centering
\setlength{\tabcolsep}{5pt}
\scriptsize
\begin{tabular}{lccc}
\hline
\textbf{Method} &
\textbf{Params (M)} &
\textbf{GFLOPs} &
\textbf{Train Time/Epoch (s)} \\
\hline
Baseline \cite{Models_Genesis} & 7.6 & 769.871 & 652.76 \\
AAGN \cite{jiang2024anatomy}   & 7.6 & 769.871 & 690.76 \\
Ours                            & 7.7 & 769.871 & 2012.14 \\
\hline
\end{tabular}
\end{table}
\section{Conclusion}
This article introduced a morphology-driven SDG framework for AD classification from 3D sMRI by combining learnable class-conditioned morphological augmentation with a 3D U-Net encoder and supervised contrastive learning. Across the external OASIS-3, ADNI, and AIBL cohorts, the proposed method achieved numerically higher mean F1 than the baseline, AAGN, and FM-SDG. After Holm correction, statistically significant improvements were observed over the baseline on ADNI and OASIS-3, and over AAGN on ADNI. Although comparisons with FM-SDG were not statistically significant, the numerical results suggest potential benefits of class-conditioned morphological augmentation for cross-cohort generalization, with the magnitude of improvement varying across cohorts. Future work will investigate computational efficiency, refinement of the augmentation strategy, and evaluation on broader datasets.

\bibliography{references}  
\end{document}